\documentclass[a4paper,12pt]{article}

 \usepackage[table,dvipsnames]{xcolor}
\usepackage{graphicx}
\usepackage{url,lineno,microtype}
\usepackage{hyperref}
\usepackage{amsmath} 
\usepackage{amssymb}  
\usepackage{amsbsy}
\usepackage{overpic}
\usepackage{todonotes}
\usepackage{verbatim}
\usepackage{subcaption}
\usepackage{booktabs}
\usepackage{tikz}
\usepackage{pgfplots}
\usepackage{dirtytalk}
\usepackage{multicol}
\usepackage[acronym,nomain,nonumberlist,nopostdot]{glossaries}
\usepackage{xcolor}
\usepackage{bbding}
\usepackage{longtable}
\usepackage[affil-it]{authblk}
 \usepackage{array,multirow,graphicx}

\newcommand{\change}[1]{\textcolor{black}{#1}}

\def\etal{\textit{et al.~}}
\def\ie{\textit{i.e.~}}
\def\eg{\textit{e.g.~}}

\newacronym{dmp}{DMP}{Dynamic Movement Primitive}
\newacronym{gmm}{GMM}{Gaussian Mixture Model}
\newacronym{gmr}{GMR}{Gaussian Mixture Regression}
\newacronym{rl}{RL}{Reinforcement Learning}
\newacronym[firstplural=Partially Observable Markov Decision Processes (POMDPs)]{pomdp}{POMDP}{Partially Observable Markov Decision Process}
\newacronym{promp}{ProMP}{Probabilistic Movement Primitive}
\newacronym{tcp}{TCP}{Tool Center Point}
\newacronym{tsp}{TSP}{Travelling Salesman Problem}
\newacronym{lfd}{LfD}{Learning from Demonstration}
\newacronym{pto}{PTO}{Projection-based Trajectory Optimization}
\newacronym{em}{EM}{Expectation-Maximization}
\newacronym{tshix}{TSHIX}{Trajectory Sampling from Human-Inspired  Exploration}
\newacronym{ftsensor}{F/T sensor}{Force-Torque sensor}
\newacronym{vsa}{VSA}{Variable Stiffness Actuator}
\newacronym{via}{VIA}{Variable Impedance Actuator}
\newacronym[firstplural=Degrees of Freedom (DoFs)]{dof}{DoF}{Degree of Freedom}
\newacronym{coc}{CoC}{Center of Compliance}
\newacronym{hri}{HRI}{Human-Robot Interaction}
\newacronym{seds}{SEDS}{Stable Estimator of Dynamical Systems}
\newacronym{lmc}{LMC}{Linear Motion with Compliance}
\newacronym{nmb}{NMB}{Nonlinear Model-Based}
\newacronym{vdc}{VDC}{Virtual Decomposition Control}
\newacronym{cmp}{CMP}{Compliant Movement Primitive}
\newacronym{mr}{MR}{Mixed Reality}
\newacronym{vr}{VR}{Virtual Reality}
\newacronym{ar}{AR}{Augmented Reality}
\newacronym{sme}{SME}{Small and Medium-sized Enterprise}
\newacronym{rbf}{RBF}{Radial Basis Function}
\newacronym{nn}{NN}{Neural Network}
\newacronym{dtw}{DTW}{Dynamic Time Warping}
\newacronym{hsmm}{HSMM}{Hidden Semi-Markov Models}
\newacronym{pca}{PCA}{Principal Component Analysis}
\newacronym{hmm}{HMM}{Hidden Markov Model}
\newacronym{zvc}{ZVC}{Zero-Velocity Crossing}
\newacronym{bic}{BIC}{Bayesian Information Criterion}
\newacronym{aic}{AIC}{Akaike Information Criterion}
\newacronym{pi}{PI}{Proportional-Integral}
\newacronym{pid}{PID}{Proportional-Integral-Derivative}
\newacronym{fri}{FRI}{Fast Research Interface}
\newacronym{rcc}{RCC}{Remote Center of Compliance}
\newacronym{ds}{DS}{Dynamical System}
\newacronym{dl}{DL}{Deep Learning}
\newacronym{cs}{CS}{Contact State}
\newacronym{mdp}{MDP}{Markov Deccision Process}
\newacronym{irl}{IRL}{Inverse Reinforcement Learning}
\newacronym{gail}{GAIL}{Generative Adversarial Imitation Learning}
\newacronym{kmp}{KMP}{Kernelized Movement Primitives}
\newacronym{umic}{UMIC}{Unified Motion and variable Impedance Control}
\newacronym{gps}{GPS}{Guided Policy Search}
\newacronym{svm}{SVM}{Support Vector Machines}
\newacronym{pbd}{PbD}{Programming by Demonstration}
\newacronym{sscp}{SSCP}{Surface-Surface Contact Primitive}

\author{Markku Suomalainen$^{1*}$, Yiannis Karayiannidis$^{2}$ and Ville Kyrki\,$^{3}$
\thanks{$^{1}$Corresponding author; Center of Ubiquitous Computing, Faculty of Information Technology and Electrical Engineering, University of Oulu, Finland {\tt\small (e-mail: markku.suomalainen@oulu.fi).}} 
\thanks{$^{2}$Systems and Control, Department of Electrical Engineering,  Chalmers University of Technology, Gothenburg, Sweden {\tt\small (e-mail: yiannis@chalmers.se).}} 
\thanks{$^{3}$Intelligent Robotics research group, Department of Electrical Engineering and Automation, School of Electrical Engineering, Aalto University, Helsinki, Finland {\tt\small (e-mail: ville.kyrki@aalto.fi).}} 
\thanks{This work was supported by Academy of Finland project PERCEPT 322637 and by European Research Council project ILLUSIVE 101020977,}
}

\begin{document}

\title{A Survey of Robot Manipulation in Contact}

\maketitle

\begin{abstract}

In this survey, we present the current status on robots performing manipulation tasks that require varying contact with the environment, such that the robot must either implicitly or explicitly control the contact force with the environment to complete the task. Robots can perform more and more manipulation tasks that are still done by humans, and there is a growing number of publications on the topics of 1) performing tasks that always require contact and 2) mitigating uncertainty by leveraging the environment in tasks that, under perfect information, could be performed without contact. The recent trends have seen robots perform tasks earlier left for humans, such as massage, and in the classical tasks, such as peg-in-hole, there is a more efficient generalization to other similar tasks, better error tolerance, and faster planning or learning of the tasks. Thus, in this survey we cover the current stage of robots performing such tasks, starting from surveying all the different in-contact tasks robots can perform, observing how these tasks are controlled and represented, and finally presenting the learning and planning of the skills required to complete these tasks.

\end{abstract}


\glsresetall
\section{Introduction}
\label{intro}
For a long time contact with the environment in robotic manipulation was considered problematic and tedious to manage. However, this paradigm is under rapid change at the moment. Tasks that have been considered too difficult for robots due to the need for delicate modulation of forces by humans are being performed by robots. Additionally, in many tasks contact is not only managed but exploited such that robots can localize themselves and their tools with the help of contact to perform tasks when faced with uncertainties. The methods for managing and exploiting contacts have also evolved such that the computational burden is not infeasible. Several workshops in the main robotic conferences during recent years have been organized under the theme of managing contact, and terms such as ''manipulation with the environment" have been coined. This is expected evolution, as humans take extensive advantage of the environment during various manipulation tasks with limited clearances \cite{klingbeil2016experimental}.

In this survey we present an overview of how robots perform contact-rich tasks, which we call \textit{manipulation in contact}; the contact can be between the robot's hand and the environment, or between a tool held by the robot and the environment. Such tasks require \textit{explicit or implicit control of the interaction forces}; either the robot or a tool held by the robot is in contact with the environment for considerable periods of time. The tasks range from traditional assembly tasks, such as screwing and peg-in-hole, to material-removing tasks such as excavation and wood planing, and even to tasks mainly performed by humans currently, such as melon scooping and massaging; in other words, tasks where the robot is in contact with the environment for an extended duration while needing to manage varying contacts with the environment. Especially in \glspl{sme} there is an increasing interest in the use of cobots, \ie robots that can exhibit compliant behavior in such tasks \cite{cencen2018design}. A more comprehensive list of such tasks is presented in Section~\ref{sec:skills}. Dynamic tasks, such as throwing or batting that are more dependent on kinematics due to the short contact time, are not considered in this survey; see~\cite{ruggiero2018nonprehensile} for a recent survey on robots performing these kinds of tasks. 

The first requirement for a robot to perform tasks while in continuous contact with the environment is a suitable low-level \textit{controller}. The first choice is between \textit{explicit} or \textit{implicit} control of the forces. For explicit control where a desired force level is set, the classical force controller, or often a hybrid force/position controller \cite{raibert1981hybrid}, is a common choice; in the classical version, a PI-controller with high-frequency updates attempts to keep the force applied by the robot at a desired level. 


Implicit control of contact forces is often referred to as \textit{compliance}. For creating compliance through software, impedance control \cite{hogan1987stable} is a popular choice, where deviation from a desired trajectory is allowed, allowing both free space and in-contact motions without switching the controller. Compliance can also be created with a mechanical device, such as a \gls{vsa} or the relatively novel field of soft robotics \cite{chin2020machine}; in this survey, we will not go deeper into mechanical compliance, but present certain works where mechanical compliance is used. More advanced versions of impedance controllers have been designed, from which we will present several novel ones in Section~\ref{sec:control}; a survey focusing only on the controllers was published six years ago \cite{khan2014compliance}. 

\begin{figure}[tbp]
	\centering
	\includegraphics[width=0.9\columnwidth]{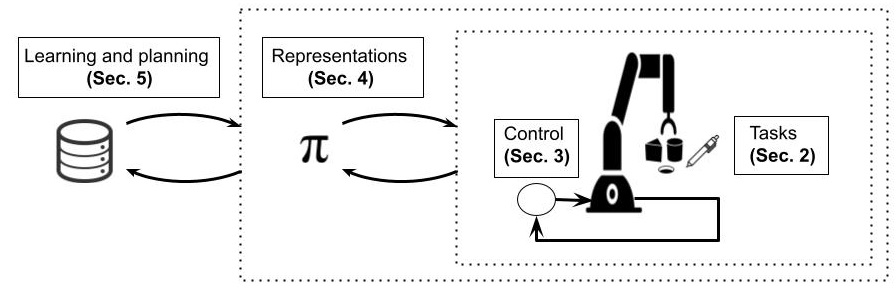}
	\caption{\change{The structure of the paper: we first present the in-contact manipulation tasks performed by robots in Section~\ref{sec:skills}, and then proceed with controllers used for in-contact manipulation in Section~\ref{sec:control}. Then, we present different ways to represent the policy $\Pi$ in Section~\ref{sec:representing}, and finally show methods for planning or learning the tasks using the representations in Section~\ref{sec:learnplan}.} }
	\label{fig:overview}
\end{figure}

Whereas sometimes the whole skill is represented at the control level, often a higher level representation is used which then passes commands to the controller. There are many ways to form these representations, and they may be hierarchical even beyond the control layer: we will survey the representations used with in-contact tasks in Section~\ref{sec:representing}. Then, in Section~\ref{sec:learnplan} we make a coarse division into three different categories the skill to perform a task can be conveyed to the robot; \textit{planning}, \textit{\gls{lfd}} and \textit{\gls{rl}}. Planning, or motion planning, is the classic method for planning the motion of a robot from known information while avoiding obstacles. Whereas exact planning for in-contact tasks is difficult, there are various modern methods that can achieve this that will be presented in this survey. In \gls{lfd} the underlying assumption is that a human user can perform the skill efficiently, and that transferring the skill to a robot results in successful and efficient execution. Finally, \gls{rl} is currently a highly popular machine learning method, where the algorithm actively searches for a good solution. An overview of the structure of the paper is presented in Fig.~\ref{fig:overview}

There are a few other related surveys which may interest the readers of this paper. For a more general survey on \gls{lfd} for other sort of skills besides in-contact skills, we recommend ~\cite{argall2009survey,osa2018algorithmic,ravichandar2020recent}; similarly, there are general surveys on \gls{rl} for robotics (\eg \cite{kober2013reinforcement}) and a very general survey on all kinds of robot learning \cite{kroemer2019review}. There are also more detailed surveys on \gls{lfd} and \gls{rl} focusing on assembly tasks \cite{xu2019compare,zhu2018robot,braun2020incorporation}. Finally, when considering vision in the loop, there is a survey of the interplay between vision and touch \cite{bohg2017interactive}; for this reason, this survey will not consider vision-based manipulation either, and we will focus on works where the sole feedback, if any, is force. Even though many frameworks could be extended to, or used on, in-contact manipulation, in this survey we focus on publications that explicitly show performance on an in-contact task.

\section{Tasks requiring manipulation in contact}
\label{sec:skills}
\begin{figure}%
  \centering
  \vspace*{-3cm}
    \begin{subfigure}{0.43\columnwidth}
        \includegraphics[width=\columnwidth]{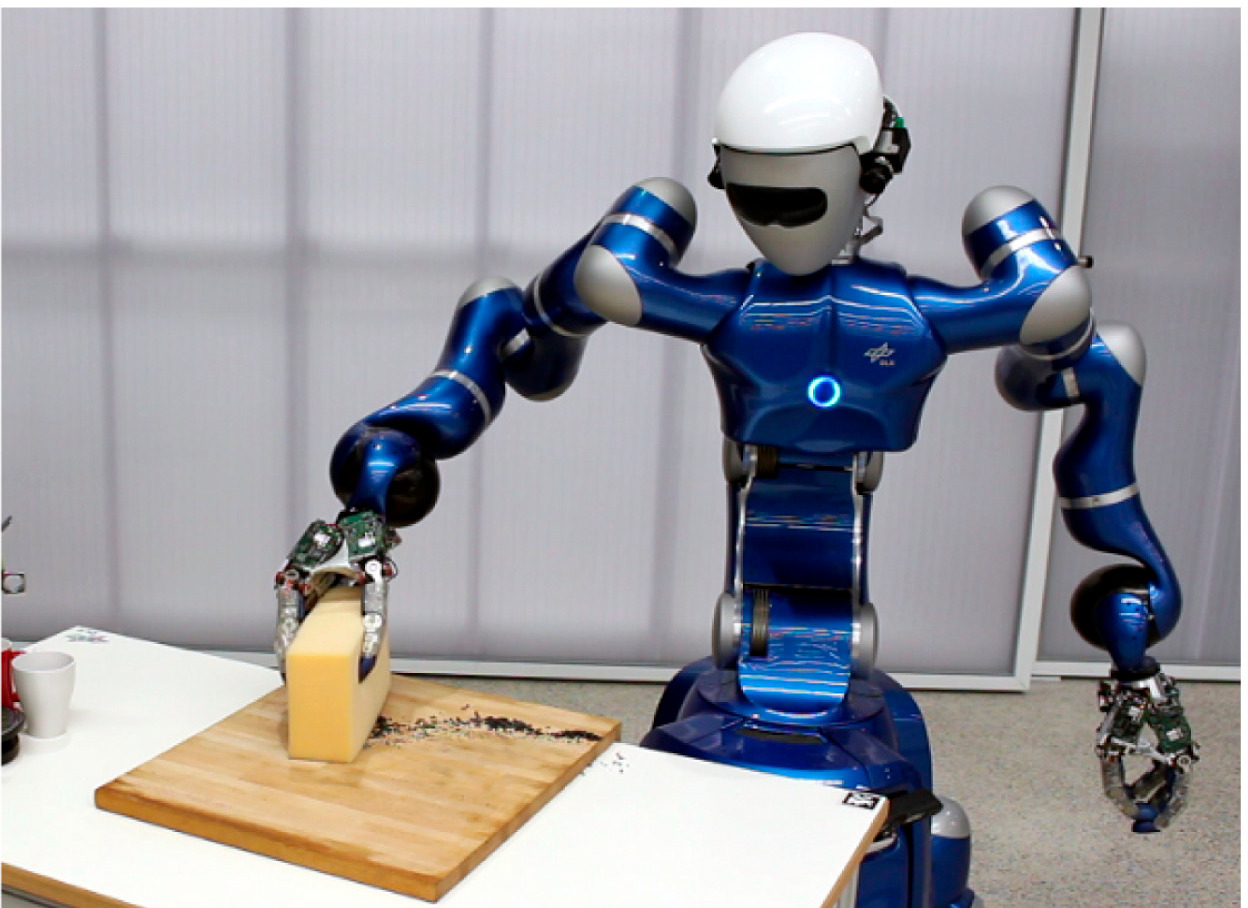}
        \caption{Wiping \cite{leidner2019cognition}}
    \end{subfigure}\hfill
    \begin{subfigure}{0.23\columnwidth}
        \includegraphics[width=\columnwidth]{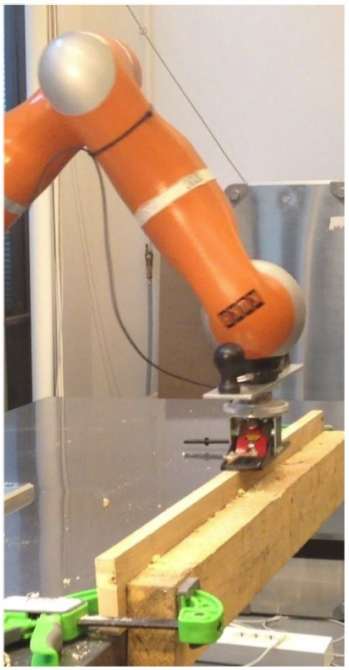}
        \caption{Wood planing \cite{montebelli2015handing}}
    \end{subfigure}\hfill
    \begin{subfigure}{0.43\columnwidth}
        \includegraphics[width=\columnwidth]{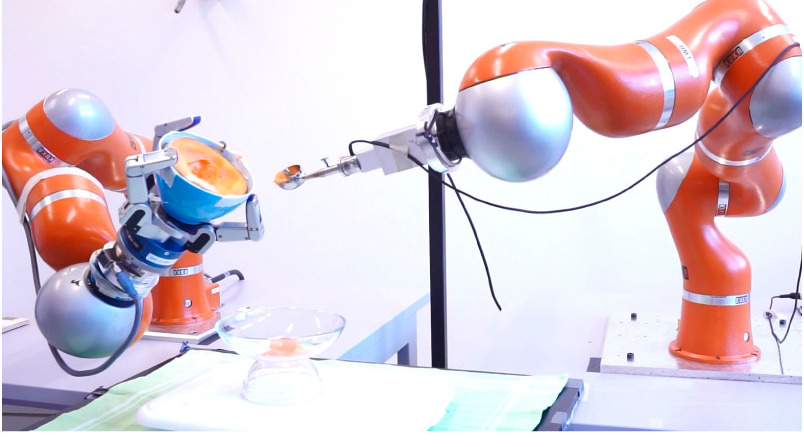}
        \caption{Scooping \cite{ureche2018constraints}}
    \end{subfigure}\hfill
    \begin{subfigure}{0.43\columnwidth}
        \includegraphics[width=\columnwidth]{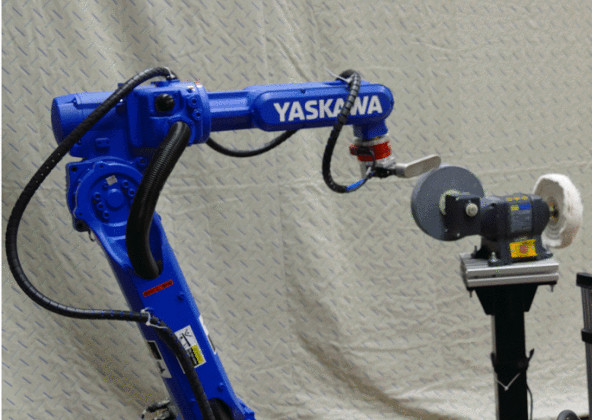}
        \caption{Grinding \cite{nemec2021virtual}}
    \end{subfigure}\hfill
    \begin{subfigure}{0.23\columnwidth}
        \includegraphics[width=\columnwidth]{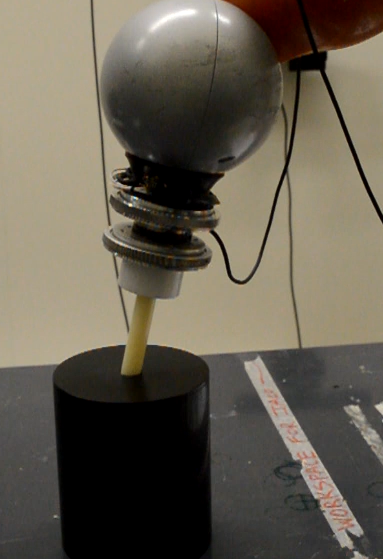}
        \caption{Classic (rounded) Peg-in-hole \cite{suomalainen2021imitation}}
    \end{subfigure}\hfill
    \begin{subfigure}{0.43\columnwidth}
        \includegraphics[width=\columnwidth]{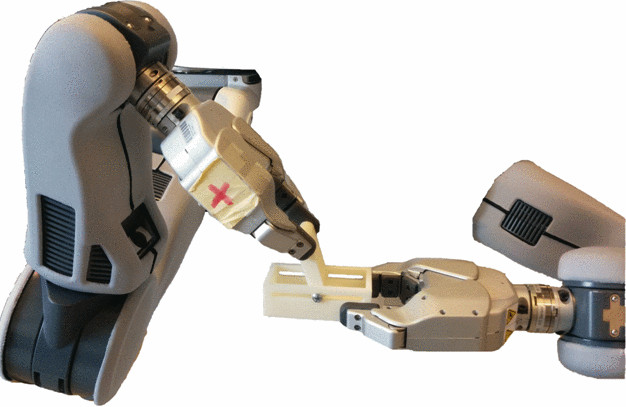}
        \caption{Folding assembly \cite{almeida2016folding}}
    \end{subfigure}\hfill
    \begin{subfigure}{0.43\columnwidth}
        \includegraphics[width=\columnwidth]{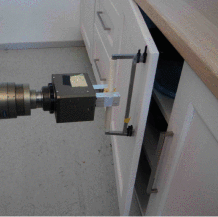}
        \caption{Door opening \cite{karayiannidis2016adaptive}}
    \end{subfigure}\hfill
    \begin{subfigure}{0.43\columnwidth}
        \includegraphics[width=\columnwidth]{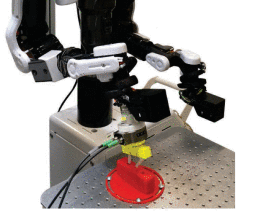}
        \caption{Multi peg-in-hole \cite{hoppe2019planning}}
    \end{subfigure}\hfill
    \caption{Different in-contact tasks done by robots}
    \label{fig:tasks}
\end{figure}

We define manipulation in contact as tasks where explicit or implicit control of the interaction forces is required. There are actually very few tasks where explicit control of the forces is strictly required; an example of such a task is tightening a screw with a specific torque value. However, even though many tasks such as polishing could be done with implicit control of the forces as well, using explicit force values may improve the success of a learned or planned task. Finally, there are tasks which could be performed without any control of the forces under perfect knowledge, such as the classical peg-in-hole and similar workpiece alignment tasks and articulated motions, such as opening a door. However, any uncertainty in such tasks raises the need for controlling the contact forces to prevent excessive collisions; moreover, by leveraging compliance a robot can perform, for example, a peg-in-hole task with clearance smaller than the robot's accuracy \cite{gubbi2020imitation}. In this section, more details of the manipulation skills requiring controlling of force interactions will be given under three categories; environment shaping, workpiece alignment, and articulated motions. Table~\ref{tab:tasks} provides a list of tasks used in the included papers where the task was a focal point of the publication, \change{and Fig.~\ref{fig:tasks} illustrates several tasks from the list}. More details on the control-related terminology used in this section can be found in Section~\ref{sec:control}.

\begin{table*}[ht]
        \centering
        \small \caption{The most popular tasks accomplished by the surveyed papers.}
        \vspace{-1mm}
        \begin{tabular}{@{}c|c @{}}
        Wiping or polishing & \cite{urbanek2004learning,deng2016learning,khansari2016adaptive,gams2016adaptation,kramberger2018passivity,leidner2019cognition,brunete2016user,qian2019sensorless,amanhoud2019dynamical,amanhoud2020force} \\ \hline  
         Grinding or similar & \cite{chebotar2014learning,hazara2016reinforcement,montebelli2015handing,steinmetz2015simultaneous,hsu2000intelligent,maric2020collaborative,ng2014method,ng2016programming,nemec2018learning,nemec2021virtual,zhang2020design} \\ \hline  
         Scooping & \cite{ureche2018constraints,kramberger2020adapting} \\ \hline
         Peg-in-hole variants & \cite{ehlers2018search,apolinarska2021robotic,hou2020fuzzy,zhang2017peg,inoue2017deep,schoettler2020meta,gubbi2020imitation,suomalainen2019dual,suomalainen2021imitation,hagos2018,abu2015adaptation,kramberger2016transfer} \\ {} & \cite{kramberger2017generalization,abu2014solving,su2018learning,hoppe2019planning,wang1996derivation,scherzinger2019contact,meeussen2007contact,su2016learning,van2018comparative,beltran2020learning,nottensteiner2020robust,jasim2014position,newman2001interpretation,park2020compliant} \\ {} & \cite{nemec2020learning,wu2020learning,khader2020stability,shao2020learning,luo2019reinforcement,schoettler2020meta,kaspar2020sim,ma2021efficient,oikawa2021reinforcement,wirnshofer2019robust,pervez2017learning} \\ \hline
         Articulated motions & \cite{almeida2016folding,stolt2011force,hayami2021error,zollner2004programming,carrera2015learning,tanwani2016learning,Niemeyer97,Lutscher12,Lutscher10,karayiannidis2012open,karayiannidis2013model,karayiannidis2016adaptive,Nemec17} \\ \hline
         Hydraulic & \cite{dadhich2015survey,maeda2014iterative,jud2017planning,dobson2016admittance,marshall2008toward,egli2020towards,suomalainen2018hydraulic} \\ \hline  
         Rare tasks & Massage \cite{khoramshahi2020arm}, velcro peeling \cite{yuan2021multi}  engraving \cite{koropouli2011learning}  
  \end{tabular}
  \label{tab:tasks}
  \vspace{-4mm}
\end{table*}

Many tasks where manipulation in contact is required also involve tools. They can either be rigidly attached to the robot arm or can be grasped by the robot for use; in this survey, we do not differentiate between these cases, except by noting that grasping a tool always creates uncertainty regarding the location of the tooltip, which increases the need for compliance when in contact. There are of course methods to alleviate this uncertainty if enough information about the tool has been properly measured (for example, \cite{karayiannidis2014online}).

We also note that there are other uses for compliance in robotics, which are however not covered in this survey; often a key point of these methods is to use a variance of trajectories as a signal when compliance is needed, by assuming that a low variance across multiple demonstrations at a stage of the demonstrations means that that location is important and thus compliance should not be allowed. Vice versa, when the variance is high, the exact location is not important and compliance can be increased too (see, for example,  \cite{calinon2010learning,rozo2013learning, kronander2014stiffness,abu2018force,kronander2012online}). However, this idea is often not applicable to in-contact tasks, since during contact variance in location is low but compliance is still required. These methods work well in  learning collaborative tasks \cite{calinon2010learning,rozo2013learning}, physical \gls{hri} \cite{gasparri2016robust,gribovskaya2011motion,nemec2018human} and in free space tasks such as delivering and pouring a drink \cite{kronander2012online,kronander2014stiffness}. 

\subsection{Environment shaping}

Shaping the environment means often removing a small layer of material either uniformly or from a fixed location. These sorts of tasks can be differentiated by the accuracy required; whereas surface wiping (removing the dust) is essentially only a surface following task, engraving and wood planing require such high precision that they are likely not possible to complete with only implicit control of the interaction force. Another way to differentiate between these sorts of skills is whether they are periodic or not: wiping can be considered a periodic task, a property exploited by some of the presented methods, whereas engraving is performed on a very detailed and fixed path. 

The simplest task in this region is perhaps wiping \cite{urbanek2004learning}, which has been revisited upon multiple papers with different methods \cite{deng2016learning,khansari2016adaptive,gams2016adaptation,kramberger2018passivity,leidner2019cognition}; in this task, the original material is not affected, since wiping mostly refers to cleaning the material. Another term used for, practically, the same task, is polishing \cite{brunete2016user,qian2019sensorless,amanhoud2019dynamical,amanhoud2020force}. These tasks have been completed with either implicit or explicit control of the forces; implicit is often enough since an important part is keeping contact with the surface. Interestingly, from the aforementioned publications, only \cite{brunete2016user,amanhoud2019dynamical} use implicit control, whereas the other works explicitly control the contact force (possibly with admittance control to allow room for error). A possible reason is that with implicit control, if there is no force feedback loop, it is possible to lose contact without noticing, but with force feedback in place, the use of explicit control emerges naturally. Finally, loosely connected to these types of tasks is massage \cite{khoramshahi2020arm}, which however requires more accurate force control and surface adaptation. 

The task gets more complicated when removing the environmental material is considered; there are multiple variations of removing the outer layer of material using a tool to make the surface smoother. These tasks are often found in industry and carpentry: scraping \cite{chebotar2014learning}, wood planing \cite{hazara2016reinforcement,montebelli2015handing,steinmetz2015simultaneous}, deburring \cite{hsu2000intelligent} and sanding \cite{maric2020collaborative}. All of these tasks employ explicit force control, as expected, and can be considered periodic; a non-periodic similar task is engraving \cite{koropouli2011learning}. Additionally, interesting related tasks are grating \cite{ureche2015task}, grinding \cite{ng2014method,ng2016programming,nemec2021virtual,zhang2020design}  and drawing \cite{steinmetz2015simultaneous, lee2013relative}, where material is removed from the tool being held instead of the environment; as expected, grinding and grating are more periodic and require more accurate force management, whereas while drawing keeping the contact is more important but there is no periodicity. In scooping often a larger amount of a softer material is removed \cite{ureche2018constraints}, but also parts for an assembly task can be scooped \cite{kramberger2020adapting}. Also, the use of a drill by a robot requires careful control of the environment forces \cite{peternel2017method,babarahmati2021robust}. Finally, a fundamentally different material removing task is velcro peeling \cite{yuan2021multi}, which has the potential to spawn other similar works for robots. 

Whereas the above examples were tasks typically performed on a human scale, similar motions performed by much larger machines can be found from earthmoving, mostly performed by excavators or wheel loaders \cite{dadhich2015survey}. The main difference is the use of hydraulics in most heavy machines, which, due to the nontrivial closed-loop kinematics and interactions of the hydraulic valves, makes controlling the forces more challenging; additionally, the material being excavated often has high variance in resistance (for example, rocks in sand). For the previously mentioned reasons, excavation is often done even without force feedback; as such, autonomous excavation has been accomplished with real machinery \cite{maeda2014iterative,jud2017planning,dobson2016admittance,marshall2008toward,egli2020towards}. However, there is also work towards estimating the forces \cite{koivumaki2015high} and using an impedance controller with hydraulic manipulators \cite{Koivumaki_TMECH2017,suomalainen2018hydraulic}. 

\subsection{Workpiece alignment}
With the term workpiece alignment we mean mainly tasks found in industrial assembly, often variations of the classic peg-in-hole; however, similar tasks are often found inside homes, such as plugging in a socket \cite{ehlers2018search} or assembling furniture \cite{apolinarska2021robotic}. There are almost infinite variations of peg-in-hole, starting with the difference in clearance (industrial assembly often requires very low clearances) and ranging to multi-peg-in-hole  where a plug which has two or three pegs that have to slide in simultaneously (for example, \cite{hou2020fuzzy}). Another flavor is dual-arm peg-in-hole; however, in this case, often one arm performs most of the motions, both with humans and robots \cite{zhang2017peg}. The mechanics, forces, and errors in peg-in-hole assembly, to enable finding the best way to perform it, have been researched from the 1980s \cite{whitney1982quasi} and are still under active research \cite{chernyakhovskaya2020peg}.

Compliance has been found to be a useful property in performing peg-in-hole type assembly motions already in the 1980s \cite{mason1981compliance} and similar research on peg-in-hole compliance is still ongoing \cite{park2020compliant}. 
In the '90s research on compliant assembly was centered around \textit{fixtures}, devices for holding workpieces during assembly. Schimmels and Peschkin \cite{schimmels1991force} showed how to insert a workpiece into a fixture guided by contact forces alone, and the topic has later been revisited in (at least) \cite{newman2001interpretation}. Yu \etal \cite{yu1995fixture,yu1998complete} then formalized the contact force guidance into a planar sensor-based compliant motion planning problem, \ie a preimage planning problem \cite{lozano1984automatic}; these methods have thrived also due to increasing accuracy of contact state estimation \cite{de1999estimating}, which is still seeing vast improvements \cite{wirnshofer2019state}. In recent works, both explicit and implicit force controls have been used; even though explicit force control may not be strictly necessary, it is often beneficial, especially in certain variants of the task. 

Regardless of the long history, a human can still outperform a robot in peg-in-hole tasks in certain metrics, such as generalization, managing surprising situations and uncertainties and really tight clearances \cite{inoue2017deep}; however, there is work to overcome these, such as meta-reinforcement learning for generalization \cite{schoettler2020meta} and clearances smaller than the robot's accuracy ($6\mu m$) \cite{gubbi2020imitation}. There are also more difficult variants, such as multi-peg-in-hole \cite{hou2020fuzzy}, coupler alignment and interlocking \cite{hagos2018,suomalainen2021imitation}, assembly construction \cite{belousov2022robotic}, hole-in-peg with threaded parts \cite{Salem20} or peg-in-hole combined with articulated motions, which include tasks such as folding \cite{almeida2016folding} or snapping \cite{stolt2011force,hayami2021error} where either more elaborate motions or force over a certain threshold is required to complete the task. Also, most works in the field assume rigid pieces, but there is also work towards the more challenging field of elastic pieces \cite{wirnshofer2019robust}. Since the peg-in-hole is such a standard problem in industry and homes, active research on the topic continues as methods and hardware evolve. For example, preimage planning \cite{lozano1984automatic} was long considered infeasible for planning compliant motions in contact \cite{canny1987new}, but nowadays it is possible to plan near-optimal plans nonetheless \cite{guan2018efficient}; similarly, there are better simulation tools under development which can greatly ease transfer learning \cite{ajay2018augmenting}. 

\change{Finally, with an increasing number of methods proposed for the peg-in-hole problem, novel benchmarks are being proposed to allow comparison experiments between methods in peg-in-hole. A classic mechanical system for assembly, the so-called Cranfield benchmark, was proposed already in 1985 \cite{collins1985development}, which includes different kinds of peg-in-hole variants to finish. However, this proposed benchmark was mainly about the mechanical structure, and there are two recent proposals for more complete benchmarks to be used to compare peg-in-hole algorithms; Van Wyk \etal \cite{van2018comparative} proposed another peg-in-hole mechanical system along with metrics such as success probability and variance to measure the success of an algorithm, and Kimble \etal \cite{kimble2020benchmarking} proposed similar metrics for small-part assembly.}


\subsection{Articulated motions}

In articulated motions the objects can only move along a pre-defined path, which may not be obvious from a visual inspection of the object; typically the robot needs to interact with  the articulated objects in order to perceive the underlying kinematic structure.
The majority of everyday articulated objects that afford single arm manipulations can be modeled as one degree of freedom mechanisms (doors, drawers, handles) even though doors with handles can be seen as a combined two degrees of freedom mechanism.  Also many opening tasks, such as jars \cite{zollner2004programming} and valve turnings \cite{carrera2015learning,tanwani2016learning}, are articulated motions. 

A level of force control is often used, although a direct force control scheme would not be strictly necessary. A big portion of the modern approaches in force-driven door opening follows the seminal work of Niemeyer and Slotine \cite{Niemeyer97} that is based on: i) the implementation of a damped desired behavior along the allowed direction of motion, and ii) the estimation of the direction of motion.  The realization of the desired behavior has been achieved through indirect compliant control \cite{Lutscher12}. A richer desired impedance has been used instead of a simple damping term in \cite{Lutscher10} realized in an admittance control framework. Alternatively, hybrid force/motion control with an implicit realization allow the use of velocity-controlled robots \cite{karayiannidis2012open,karayiannidis2013model,karayiannidis2016adaptive}. While traditionally only the linear motion is considered, in the last decade orientation control is also studied both in the context of adaptive \cite{karayiannidis2016adaptive} and learning control \cite{Nemec17}.

The estimation of the direction of motion while opening a door/drawer is a specific case of estimation of constrained motion \cite{DeSchutter88,Yoshikawa93}. Velocity-based (twist-based) estimation has been employed aiming at reducing chattering due to
measurement noise and dealing with the ill-definedness of the normalization
for slow end-effector motion such as spatial filtering \cite{Niemeyer97}, moving average
filters \cite{Lutscher10} employing simple drop-out heuristics \cite{Lutscher12}. From a control perspective, decoupled estimation and control is considered to be indirect adaptive control as opposed to direct adaptive control approaches \cite{karayiannidis2016adaptive} that are more robust, particularly for cases where the estimation dynamics suffer from lags. The reader is referred also to  \cite{karayiannidis2016adaptive} for a richer summary of the door opening literature until 2015. While the estimation of motion directions for simple one-degree mechanisms is possible using proprioception and force measurements, the tracking of multi-body articulated objects requires richer perception input including vision and probabilistic methods to deal with noise and uncertainty in the models, e.g. when the contact is lost \cite{Martin2021}.


\subsection{Dual-arm manipulation in contact}
A survey focusing on dual-arm manipulation can be found in \cite{smith2012dual}, where the focus is mainly on centralized settings with two arms that can get feedback from each other. We will not focus on the specifics of dual-arm systems in this paper, but certain methods presented are also applicable to dual-arm methods. Thus, we will shorty consider performing tasks with two arms instead of one, and in the remainder of this survey dual-arm tasks are treated together along with single-arm tasks.

Considering industrial assembly tasks, \cite{yamada1995development} did a thorough analysis of industrial assembly tasks occurring within Toyota car manufacturing, but did not report cases where simultaneous motions of two arms were required; thus, the second arm in many tasks acts as a fixture. However, in a classic articulated rotation task such as opening a jar, rotating a pepper mill, or just manually tightening nuts and bolts \cite{zollner2004programming,kroemer2014learning,liza2011assembly} the task may be performed faster if both arms move in harmony, reducing the number of required grip changes. Also in tasks such as melon scooping while one arm holds and the other one scoops \cite{ureche2018constraints}, drawing on a board held by another robot \cite{lee2013relative} or peg-in-hole where also the ''hole" is held \cite{zhang2017peg,suomalainen2019dual}, it can be advantageous if also the holder arm can explicit compliance, even if no active motions are performed; this is also how a human performs such tasks \cite{zhang2017peg}.
Dual-arm manipulation has been employed for manipulating 2-DOF articulated objects with co-located prismatic and revolute joints where F/T sensing and proprioception of both were utilized to localize the joints and track their states \cite{Almeida18}; such an approach can be used for modeling and performing bimanual folding assembly.

\section{Control of manipulation in contact}
\label{sec:control}
Robots perform complex tasks often with a hierarchical policy where control is the lowest level; essentially, it translates the motion or force commands to actual motor currents on a high frequency. In this survey, we will focus more on the higher-level policies, but nonetheless introduce the different control methods used together with the higher-level policies; for a survey focusing more on compliant control methods, see, \eg,~\cite{khan2014compliance}. The majority of force control approaches can be divided into the following groups: \textit{i) direct} \cite{raibert1981hybrid} and \textit{ii) indirect control} \cite{hogan1985impedance} based on the force control objective and  \textit{i) explicit} and \textit{ii) implicit} based on the nature of the input control signal. Specifically, we introduce the main control methods for controlling the forces directly (direct force control)  and for controlling a dynamic relationship between forces and displacements (impedance and admittance control) and their use in in-contact skills.





\subsection{Force Control}
\label{sec:forcecontrol}
\change{
A straightforward method for managing manipulation in contact without damaging the robot or the environment is \textit{direct force control} \cite{whitney1977force,whitney1985historical}, essentially a feedback controller trying to maintain a desired contact force. For more general use cases, force control is often combined with position control for driving the contact force to a desired set-point or trajectory (regulation/tracking problem respectively). There are at least two distinct methods for accommodating the  force and  motion control spaces in the control action (more details in Table~\ref{tab:controllers}). \textit{ Hybrid force/motion controllers} are formulated based on an explicit decomposition of control subspaces based on the motion constraints, whereas \textit{parallel force/motion control} does not use constraints in order to define different control subspaces; instead, a force control loop based on integral action dominates over a compliance-based position control loop. Another dichotomy for different force control applications depends on whether the robot is torque-controlled or not \cite{Roy2002}. As force measurements are often noisy and thus their differentiation does not produce meaningful results, usually a \gls{pi} controller is used in the force control loop and the update frequency must be high enough to avoid instabilities \cite{colgate1989analysis}. 
}

\change{
Table \ref{tab:controllers} summarizes several control schemes that can be used to control the interaction between a robotic manipulator and a contacted surface\footnote{For presentation clarity and simplicity, we consider a contact problem that does not involve control of orientations and torques}. The first three rows present three different force/position controllers for torque-controlled robots, and the next three rows present the same controllers for velocity-controlled robots. Before going into further details, it is important to observe that the compliance of the environment plays a role in whether the desired set position and force are reached. A position setpoint can be reached only if it is defined according to the compliance of the contact and the position of the plane; it is possible that putting a certain amount of pressure on the environment causes such deformation that the set point location is no longer valid. Thus, different methods for handling the interplay between the environment and the robot are required.}

\change{
To manage even deforming environments, hybrid(1st and 4th row) and parallel(2nd and 5th row) force/position control structures are minimizing the position error by achieving equilibrium $\pmb{Q}(\pmb{p}-\pmb{p}_d)=0$ without requiring knowledge of the compliance of the contact and the position of the plane. However, they both assume knowledge of the normal vector $\pmb{n}$ either for defining the projection matrices $\pmb{N}$ and $\pmb{Q}$ (hybrid force/position control) or defining a geometrically consistent desired force $\pmb{Q} \pmb{f}_d=\pmb{0}$ (parallel force/position control). Studies on uncertainties on the slope of the surface can be found for the regulation problem in \cite{TAC08,RAS07} (hybrid) and \cite{IET07} (parallel) and for the force/position tracking problem in \cite{RAS10,ICRA07} (compliant contact) and \cite{AUT09} (rigid contact.)} 

\change{
On the 3rd and 6th lines in Table \ref{tab:controllers} we present a standard 
\textit{stiffness controller}\footnote{The stiffness controller can be also considered a special case of the impedance controller presented in Section~\ref{sec:impedancecontrol} leaving out inertia and damping from (\ref{eq:impedance})} that can be used for contact and non-contact cases; whereas parallel and hybrid force controllers are not suitable for free space motions due to the desired nonzero contact force value, a stiffness controller can also perform trajectories in free space. High stiffness values make the controller resemble a position controller; this reduces the position error caused by low stiffness values, but contact with the environment can generate high, undesirable contact forces. Also, it needs to be remembered that the system's (robot and environment) total behavior depends on the ratio of stiffnesses between the robot and the environment; thus, a rigorous stability proof also requires assumptions, or knowledge, about the environment. The stiffness values for different directions, encoded in the matrix $\pmb{K}_P$, need to be carefully chosen considering the task, the environment, and all constraints, paying special attention to cases where the environment cannot be considered stiff. The stiffness control scheme does not necessarily require force measurements, but, if an implicit implementation is attempted, force measurements are required; see the (inverse) damping velocity control scheme in the sixth row of Table \ref{tab:controllers}. 
}

\begin{table}
{\footnotesize 
\begin{tabular}{|c|r |rll|l| } 
 \hline

\multirow{4}{*}{\rotatebox[origin=c]{90}{\scriptsize{~~Explicit}}}& \multirow{2}{*}{\rotatebox[origin=c]{0}{\scriptsize{Direct}}}&
        \cellcolor{blue!25}  ${\pmb{u}} = \pmb{g}(\pmb{q})-\pmb{K}_D(\pmb{q})\dot{\pmb{q}} $ &  \cellcolor{ForestGreen!25} 
          +$\pmb{J}^T\pmb{f}_d$ $-k_f\pmb{J}^T\pmb{N}(\pmb{f} - \pmb{f}_d) $ & \cellcolor{red!25} 
          $-\pmb{J}^T\pmb{Q}\pmb{K}_P \pmb{Q} (\pmb{p} - \pmb{p}_d)$ \\ 
 & &\cellcolor{blue!50} ${\pmb{u}} = \pmb{g}(\pmb{q})-\pmb{K}_D(\pmb{q})\dot{\pmb{q}} $&  \cellcolor{ForestGreen!50}  +$\pmb{J}^T\pmb{f}_d$$-\pmb{J}^T(k_f(\pmb{f} - \pmb{f}_d) +k_I \int_0^t(\pmb{f} - \pmb{f}_d))d\sigma) $ & \cellcolor{red!50} $-k_P\pmb{J}^T(\pmb{p} - \pmb{p}_d)$  \\ 
&   \multirow{1}{*}{\rotatebox[origin=c]{0}{\scriptsize{Indirect}}}& \cellcolor{blue!25}      {\multirow{1}{*}{${\pmb{u}} = \pmb{g}(\pmb{q})-\pmb{K}_D(\pmb{q})\dot{\pmb{q}} $}} &\cellcolor{ForestGreen!25}  & \cellcolor{red!25} \multirow{1}{*}{$-\pmb{J}^T(\pmb{q})\pmb{K}_P (\pmb{p} - \pmb{p}_d)$} \\ 

 \hline
\multirow{4}{*}{\rotatebox[origin=c]{90}{\scriptsize{~~Implicit}}}  & 
\multirow{2}{*}{\rotatebox[origin=c]{0}{\scriptsize{Direct}}}&
 \cellcolor{blue!25}   $\pmb{v}=$       &
 \cellcolor{ForestGreen!25}   $-k_f\pmb{J}^{-1}\pmb{N}(\pmb{f} - \pmb{f}_d) $ &  \cellcolor{red!25} 
   $-\pmb{J}^{-1}\pmb{Q}\pmb{K}_P \pmb{Q} (\pmb{p} - \pmb{p}_d)$ \\ 
& &\cellcolor{blue!50} $\pmb{v}=$ 
& \cellcolor{ForestGreen!50} $-\pmb{J}^{-1}(k_f(\pmb{f} - \pmb{f}_d) +k_I \int_0^t(\pmb{f} - \pmb{f}_d))d\sigma )$ 
& \cellcolor{red!50}$-k_P\pmb{J}^{-1}(\pmb{p} - \pmb{p}_d)$  \\ 

 &\multirow{1}{*}{\rotatebox[origin=c]{0}{\scriptsize{Indirect}}}&  \cellcolor{blue!25} {\multirow{1}{*}{$\pmb{v}=$}}& \cellcolor{ForestGreen!25}   {\multirow{1}{*}{ $\pmb{J}^{-1} \pmb{f}$} }& \cellcolor{red!25}  \multirow{1}{*}{$-\pmb{J}^{-1}\pmb{K}_P (\pmb{p} - \pmb{p}_d)$}  \\ 
 \hline

\end{tabular}
} \caption{\change{
A collection of explicit (torque-resolved $\pmb{u}$) and implicit (velocity-resolved $\pmb{v}$) force/position controllers: hybrid (rows 1 and 4), parallel (2 and 5) and stiffness (3 and 6) controllers, with the general control axiom highlighted in blue, force loop in green and the position loop in red. The presented controllers are meant for a non-redundant  robot with Jacobian $\pmb{J}(\pmb{q})\in \Re^{3\times3}$ (the argument $\pmb{q}$ is omitted in the expressions of the controllers to avoid notation clutter) and gravity vector $\pmb{g}(\pmb{q})\in \Re^3$ in contact with a surface. The orientation of the robot's end-effector with respect to the environment is described by a unit three-dimensional surface normal vector $\pmb{n}$  with vectors $\pmb{f}, \;\pmb{p}\in \Re^3$ describing the exerted linear force and the position of the end-effector. The desired task is encoded in the desired position set-point $\pmb{p}_d$ and desired force $\pmb{f}_d$ (direct force control) or the positive definite stiffness matrix $\pmb{K}_P$ (indirect force control). $\pmb{K}_D$ is a positive definite matrix that could depend on the configuration $\pmb{q}$ used as a gain for the joint velocities $\dot{\pmb{q}}$. When $\pmb{K}_D$ is constant the damping action is realized in joint space, when $\pmb{K}_D(\pmb{q}):=\pmb{J}^T(\pmb{q})\pmb{K}_D\pmb{J}(\pmb{q})$ the damping action is realized in the Cartesian space. The constants
$k_f$, $k_I$, $k_P$ are in general strictly positive gains but for explicit controllers $k_f$ can also be set to zero. $\pmb{N}=\pmb{n}\pmb{n}^T$ and $\pmb{Q}=\pmb{I}_3-\pmb{n}\pmb{n}^T$ are projection matrices projecting along the surface normal and tangent respectively.}}\label{tab:controllers}
\end{table}

There are several methods for measuring the force data required for hybrid and parallel force control. In Cartesian control, a straightforward method is to place a \gls{ftsensor}, which can be added afterwards to most robots, at the wrist of the manipulator. For electrically actuated robots, other methods include motor currents, motor output torques \cite{stolt2012force,qian2019sensorless} or torque sensors at the joints, with explicit force sensing providing better accuracy than the implicit methods of joint motors. For hydraulically actuated manipulators, forces from possibly hundreds of kilograms of payload can easily damage most available \glspl{ftsensor}. The method analogous to motor current estimation is using the hydraulic fluid chamber pressures for estimating the Cartesian wrench \cite{koivumaki2015stability}.




\subsection{Impedance control}
\label{sec:impedancecontrol}

\begin{figure}%
  \centering
    \begin{subfigure}{\columnwidth}
        \includegraphics[width=\columnwidth]{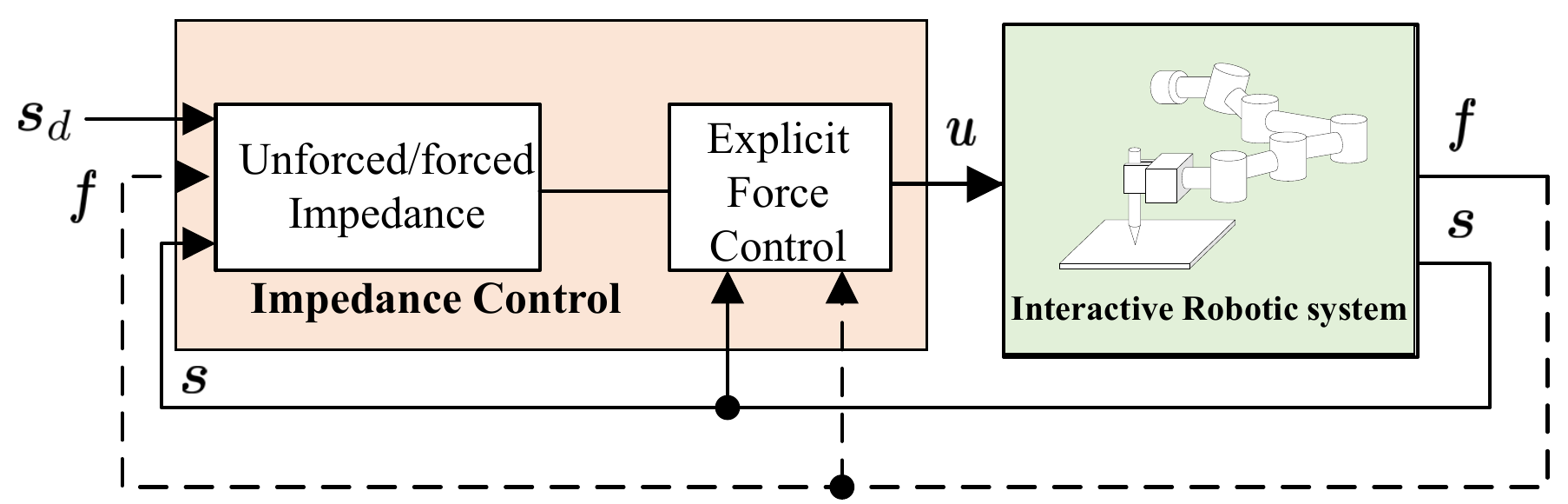}
        \caption{Impedance-controlled robotic system }
                \label{fig:impedance}
    \end{subfigure}\hfill
    \begin{subfigure}{\columnwidth}
        \includegraphics[width=\columnwidth]{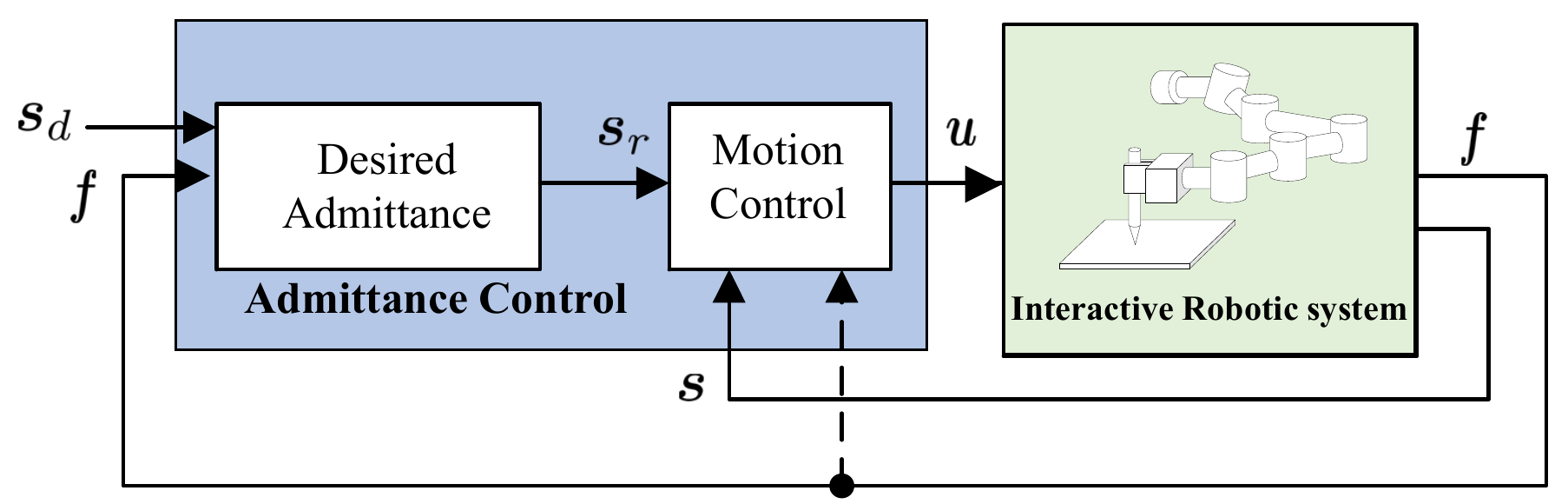}
        \caption{Admittance-controlled robotic system}
          \label{fig:admittance}
    \end{subfigure}\hfill
  \caption{Block diagrams of an impedance and an admittance controller. $\pmb{s}$, $\pmb{f}$, $\pmb{u}$ denote the state, the measured force, and the control input respectively. Both controllers take as input the desired state $\pmb{s}_d$. Note that in admittance control there is a distinct division between the filter producing the reference $\pmb{s}_r$ that may by position or velocity or both and the inner loop motion controller that generates the control input. Similar division of inner and outer control loops is also used in inverse dynamics implementations of impedance control as it is implied by the two sub-blocks in the Impedance control block (a).    }
    \label{fig:imp_adm}
\end{figure}

Impedance control is an implicit force control; the idea is to follow a trajectory in space, but allow displacements from the trajectory in proportion to set gains, in essence placing a virtual spring to the end-effector of the robot. \change{Consider a simple contact task where the robot needs to follow a desired position trajectory $\pmb{p}_d(t)$, $\dot{\pmb{p}}_d(t)$, $\ddot{\pmb{p}}_d(t)$ while possibly being in contact with the environment exerting a force $\pmb{f}$. In impedance control, instead of trying to control the force, the aim is to achieve a desired impedance in the form of a virtual spring-damper-mass system:
\begin{equation}\label{eq:impedance}
\pmb{M}(\ddot{\pmb{p}}-\ddot{\pmb{p}}_d)+\pmb{K}_D(\dot{\pmb{p}}- \dot{\pmb{p}}_d)+  \pmb{K}_P(\pmb{p}-\pmb{p}_d)=-\pmb{f}
\end{equation}
where $\pmb{M}$, $\pmb{K}_D$, $\pmb{K}_P$ are the desired apparent inertia, damping and stiffness parameters. In Fig. \ref{fig:impedance}, the basic control structure is shown; based on Eq. (\ref{eq:impedance}) and by using feedback of $\pmb{f}$ and the state $\pmb{s}$ that consists of $\pmb{p}$ and  $\dot{\pmb{p}}$, we can generate an acceleration command for an inverse dynamics controller \cite{RFC00}. When no force measurements are available, a desired apparent inertia cannot be achieved. For regulation problems, i.e. a constant $\pmb{p}_d$, a proportional controller with damping and gravity compensation without force measurements becomes equal to the stiffness controller (see Table \ref{tab:controllers}, third line) and can endow the robot end-effector with a desired tunable ``stiff'' behavior. Additionally, similar considerations regarding the stiffness of the environment are considered in impedance controller as in the stiffness controller.
} 

Whereas the original work  by Hogan \cite{hogan1987stable,hogan1985impedance} considered the impedance to remain unchanged, there are nowadays various works considering varying the impedance during execution and how that affects the stability of the system \cite{kronander2016stability,shahriari2017adapting, shahriari2019power}. Similar to a traditional spring, a virtual spring can store energy, and thus the traditional impedance controller is not passive \cite{kronander2016stability}, which can cause instabilities, especially when combined with learning approaches \cite{khader2020stability} and dealing with non-stiff environments. There are several possible solutions to this problem, for example dissipating the energy into non-relevant directions \cite{kronander2016passive}, storing it into a tank \cite{ferraguti2013tank} or filtering the impedance profile if it causes instabilities \cite{bednarczyk2020passivity}. Also, it should be noted that an impedance controller can also work as a feed-forward controller, thus avoiding the need for a force sensor or other feedback loop to simplify the robot's requirements \cite{roveda2020assembly,suomalainen2021imitation}. For complex tasks, like food cutting \cite{Mitsioni}, the linear impedance equation (\ref{eq:impedance}) cannot capture the complex interaction behaviors that are required, and instead, a desired finite-horizon cost achieved through learning-based MPC is proposed.    Finally, whereas most of the work has been developed in the context of electric manipulators, stable impedance control has also been designed for hydraulic manipulators \cite{Koivumaki_TMECH2017}. 

\subsection{Admittance control}
The idea of admittance control is to allow deviations from the commanded force in proportion to set gains, similarly to impedance control allowing displacements from the trajectory. It allows for implementing a desired impedance without necessarily relying on torque-controlled robots. With admittance control we refer here to all types of force-to-desired-motion relationships that can feed inner motion control loops, as depicted in Fig.~\ref{fig:admittance}; \change{note that the reference state $\pmb{s}_r$ (position and/or velocity) given as input to the motion control is determined by modifying the desired state $\pmb{s}_d$ using the output of differential equation (desired admittance) forced by the measured external force. The desired admittance is traditionally defined based on a linear spring-damper-mass system similar to Eq. \ref{eq:impedance} with $\pmb{p}$, $\dot{\pmb{p}}$ being however the output of the filter. For tasks involving also orientation control, the desired admittance(impedance) depends on the selection of a suitable orientation representation; recent works use dual-quaternions to encode a generalized position of the end-effector for admittance control \cite{FonsecaRAL20}.} 
Admittance control has been  been used to perform compliant motions \cite{seraji1994adaptive} and accommodation control  for force-guided robotic assembly applications \cite{Schimmels92}  for a long time already, and more lately in tasks such as excavation \cite{marshall2008toward,dobson2016admittance} and peg-in-hole \cite{kramberger2016transfer,kramberger2017generalization,kramberger2018passivity}.
%
 Admittance control has been extensively used in physical HRI \cite{sharkawy2018variable}.
In this context, recent works \cite{Keemink18,landi2017admittance} study admittance  parameter selection, propose  variable admittance structures \cite{ferraguti2019variable} or even nonlinear admittance filters  based on adaptive Dynamic Movement Primitives \cite{SidiropoulosICRA21}.

\section{Representing manipulation in contact}
\label{sec:representing}
To make a robot execute a skill to complete a task, the robot must have a representation of the skill, essentially a mapping from the task requirements and sensor feedback to controller inputs. This mapping is often called a policy, which tells the robot what action to take when receiving sensory inputs\footnote{Even though some of the surveyed methods work without force sensor feedback, at least proprioceptive feedback from, \eg, the joint encoders of a manipulator are used in practically all the applications, and thus we do not consider any of the approaches to be completely feed-forward, even if we use the term later for approaches that don't require force information. } $h$ in a state $x$, thus commanding the lower-level controller to apply certain forces or move to a certain location with an action $u$. Thus, a policy $\pi$ maps state and sensor input $h$ into actions, $u=\pi(h(x))$. In its most straightforward form, $\pi$ can be a discrete set of state-action pairs, but this is only feasible in small problems. Thus, $\pi$ is often a condensed representation of such state-action pairs. 

Policies are often hierarchical, meaning that a policy $\Pi$ first chooses to either apply policy $\pi_1$ or $\pi_2$ based on $h(x)$, and then $\pi_1$ and $\pi_2$ give the commands that the low-level controller can interpret; these \textit{continuous}\footnote{We mean continuous in the sense that the ``continuous'' policy has a wide, often continuous range of outputs (motion or force commands it can feed to the controller), whereas the ``discrete'' policy simply makes the choice between subpolicies.} representations will be covered in Section~\ref{sec:cont_rep}. Then in Section~\ref{sec:disc_rep} we will cover the discrete representations $\Pi$ which consider choosing a suitable subpolicy at each moment; we note, however, that certain methods such as \gls{hmm} and its extensions are often used as both $\Pi$ and $\pi_i$, but we present them only in Section~\ref{sec:cont_rep} for clarity. A list of the papers in this survey using certain representations can be found in Table~\ref{tab:representations}.

\begin{table*}[ht]
        \centering
        \small \caption{The most popular representation methods used in the surveyed papers.}
        \vspace{-1mm}
        \begin{tabular}{@{}c|c @{}}
       DMP \& extensions & \cite{steinmetz2015simultaneous, montebelli2015handing, hazara2016reinforcement, nemec2018learning,nemec2021virtual, deng2016learning,gams2016adaptation,kramberger2020adapting} \\ {} & \cite{kramberger2018passivity,gams2014coupling,abu2014solving,abu2015adaptation,abu2018force,denivsa2016learning,kramberger2016transfer,kramberger2017generalization,petrivc2018accelerated,tykal2016incrementally} \\ \hline 
       GMM-based & \cite{rozo2015learning,ma2021efficient,figueroa2016learning, ureche2018constraints,khoramshahi2020arm,amanhoud2019dynamical,amanhoud2020force} \\ \hline
       HMM \& extensions & \cite{kroemer2014learning,kroemer2015towards,hagos2018,figueroa2016learning,figueroa2017transform,racca2016learning,di2012hierarchical,le2021learning} \\ \hline
       (PO)MDP & \cite{beltran2020learning,calder2016wafr,guan2018efficient,Koval-ijrr2016,Sieverling2017,toussaint2014dual,de2016learning,de2013learning,de2014learning} \\ \hline
       NN & \cite{levine2015learning,arndt2021few,inoue2017deep,apolinarska2021robotic}
  \end{tabular}
  \label{tab:representations}
  \vspace{-4mm}
\end{table*}

\subsection{Continuous representations}
\label{sec:cont_rep}
Perhaps the most straightforward continuous representations are polynomials, and indeed, polynomial splines are used even in recent papers for representing robot motion policies \cite{van2017spline}. However, policies in in-contact skills often employ other kind of representations, and thus in this section we will focus on representations that have been successfully used with in-contact tasks. Many of these representations are probabilistic, often taking advantage of structures such as \glspl{gmm}, to model the uncertainty always present with in-contact tasks; often these are combined with dynamical systems to achieve time-invariant and stable behavior. Others are simply parametric function approximators, such as neural networks, or spring-damper systems built around function approximators, such as the \gls{dmp}. 
In this section we will show how these representations can be used to make an autonomous machine perform motions requiring contact with the environment. 

A very popular family of motion primitives are based on \gls{dmp} \cite{schaal2006dynamic}, employed mainly with \gls{lfd} but also with \gls{rl} especially for further improvements after a human demonstration (for example \cite{hazara2016reinforcement}). \change{A \gls{dmp} can be dissected as follows: the main component is the \textit{transformation system}, which can be written as
\begin{equation}
\kappa \ddot{\pmb{x}} = k_z (d_z (\pmb{g} - \pmb{x}) - \dot{\pmb{x}}) + f(q;y)
\label{eqt:dmp_transform}
\end{equation}
where $\kappa$ is a time constant (temporal scaling factor), $g$ is the goal and $x$, $\dot{x}$, $\ddot{x}$ are position, velocity and acceleration (the output of the system). The positive parameters $k_z$ and $d_z$ are related to damping and stiffness. This is 
a simple linear dynamical system acting as a spring-damper and driving the system forwards;  the \textit{forcing function}, $f(q;y)$,
a function approximator which perturbs the system from being a simple attractor and essentially creates the trajectory. Traditionally the forcing function is a \gls{rbf} \cite{hardy1971multiquadric}, but other options such as a \gls{nn} \cite{pervez2017learning} have been proposed. As \glspl{dmp} are learned distinctively for each \gls{dof}, either in Cartesian or joint space, the \glspl{dof} need to be synchronized, which is performed with the third component $q$, the \textit{canonical function} of which the forcing function depends, written as $\dot{q} = -\tau a_q q$, where $\tau = 1/T$ with $T$ being the length of the demonstration and $a_q$ a term managing the speed of the motions. } 

This classic \gls{dmp} was used for position only; however, there have been multiple updates on it, some of which allow \gls{dmp} to be used for in-contact tasks. The simplest way is to use \glspl{dmp} directly with explicit force control \cite{steinmetz2015simultaneous, montebelli2015handing, hazara2016reinforcement, nemec2018learning, deng2016learning,gams2016adaptation} or with admittance control \cite{kramberger2020adapting,kramberger2018passivity}. Another logical step is to augment a trajectory in space with a force or admittance profile \cite{gams2014coupling,abu2014solving,abu2015adaptation} or an impedance profile \cite{abu2018force}, which is synchronized with the canonical function $q$ as any other degree of freedom; this is sometimes called \gls{cmp} \cite{denivsa2016learning}, when the superposed torque signal is encoded with \glspl{rbf} called \textit{torque primitives}. Furthermore, \gls{cmp} has also been task-parametrized \cite{kramberger2016transfer,kramberger2017generalization} and thus generalized to new situations if the parameters of the tasks are known: for example, the depth of a hole where a peg is inserted. This means there is a direct mapping from the task parameter, such as the depth of the hole in peg-in-hole, to the weights of the primitive to easily perform variations of a similar task. Additionally, one could build a library of different \gls{cmp} skills with the task-parametrization \cite{petrivc2018accelerated}. Other modifications of \gls{dmp} include incremental learning with impedance control \cite{tykal2016incrementally}, \gls{promp} \cite{paraschos2015model} and \gls{kmp} \cite{huang2019kernelized}, which however during the writing of this survey have not explicitly been shown to perform in-contact tasks, even if capable. 

\begin{figure}[tbp]
	\centering
	\includegraphics[width=0.6\columnwidth]{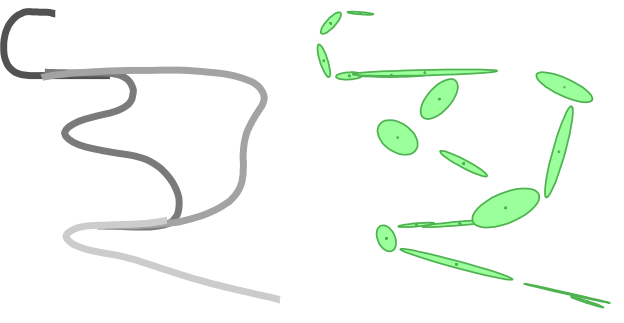}
	\caption{\change{An illustration of Gaussian distributions in a GMM (right) to encode a set of demonstrated trajectories (left) \cite{calinon2016tutorial}.} }
	\label{fig:gmm}
\end{figure}

A different approach is to use a \gls{gmm} to encode the trajectory, essentially placing a number of \change{multidimensional Gaussian distributions ${\mathcal {N}}({ \boldsymbol{\mu _{i},\Sigma _{i})}}$ along the trajectory, and learning suitable mean $\boldsymbol{\mu_i}$ and covariance $\boldsymbol{\Sigma_i}$ parameters for them; this is illustrated in Fig.~\ref{fig:gmm}.} There is a wide variety of different methods that take advantage of \glspl{gmm}, where a major difference is in the way the transitions between different Gaussians are handled. A straightforward solution is the use of \gls{gmr}, which first builds a joint distribution of the Gaussians and then derives  the regression function from the joint distribution \cite{calinon2007learning,calinon2016tutorial}. As with \gls{dmp}, the vanilla solution of \gls{gmm}-\gls{gmr} is not suitable for in-contact tasks, but augmenting it with a force profile creates the possibility of performing in-contact tasks too \cite{rozo2015learning}. Similarly as with \gls{dmp}-based methods, also \gls{gmm}-\gls{gmr} can be improved with \gls{rl} after demonstration for in-contact tasks (for example, multi-peg-in-hole \cite{ma2021efficient}). 

Another method for exploiting \gls{gmm} in a learning task is \gls{seds} \cite{khansari2011learning}. \gls{seds}, unlike \gls{dmp} and \gls{gmr}, is time-invariant due to the nature of the dynamical systems, and has also been shown to be stable with a nonlinear \gls{ds}. Whereas the original version is again for position only, there have been extensions that allow the usage in in-contact tasks. Dynamical systems have been shown to manage variable impedance control \cite{figueroa2016learning, ureche2018constraints,khoramshahi2020arm} for in-contact tasks, and also recently to provide accurate force control on unknown surfaces \cite{amanhoud2019dynamical,amanhoud2020force}. 

With the rapid rise of \gls{dl} in the recent years, \glspl{nn} have become popular as general function approximators encoding control policies, in both \gls{lfd} and \gls{rl}, potentially using human demonstrations as starting points. Methods that use \gls{dl} can be categorized based on the modeling target: the deep networks can encode either directly the control policy (e.g.\ \cite{levine2015learning}) or a trajectory for a predefined low-level controller (e.g.\ \cite{arndt2021few}). In addition, it would be also possible to use deep networks to encode environment dynamics, but this is seldom done in in-contact manipulation, most likely due to the difficulty of collecting sufficient training data for the data-hungry models. To address the need for data, other methods such as trajectory optimization can be used for generating sufficient data while the neural network will allow the use of a single policy that generalizes by interpolating the training data \cite{levine2015learning}. For situations where the policy provides an entire trajectory, the data efficiency can also be increased by using a low-dimensional latent trajectory embedding learned from unlabeled data \cite{arndt2021few}.

Finally, there are several other methods which encode in-contact tasks; another one besides \gls{seds} taking advantage of dynamical systems is \gls{umic} \cite{khansari2017learning}, which combines motion and variable impedance in a time-invariant system, showing stability of the system. Unlike the previous methods, \gls{umic} is not a trajectory following system, and requires more than a single demonstration. \gls{lmc} \cite{suomalainen2021imitation} was designed for a more narrow use case of \gls{lfd} in workpiece alignment tasks, and has also been used with hydraulic manipulators \cite{suomalainen2018hydraulic} and in a dual-arm setting \cite{suomalainen2019dual}. Another method for a special case of maintaining contact between two surfaces, \gls{sscp}, was suggested in \cite{khansari2016adaptive}. Also, there are a few methods where the impedance profile can be added to any kind of trajectory representation, learned either through \gls{lfd} \cite{abu2018force} or Bayesian optimization \cite{roveda2020assembly}.

\subsection{Discrete representations}
\label{sec:disc_rep}
Discrete representations are typically one level higher in a hierarchical policy than the continuous representations presented in the previous section; a classical example fitting to the topic at hand is to have a different policy for times before and after the contact with the environment is initiated \cite{Koval-ijrr2016} or by completely interleaving these two strategies \cite{Sieverling2017}. Another, often proposed idea, is to have a library of continuous policies (often referred to as skills) available, from which the robot can  choose a correct one at a correct time and for a correct context \cite{lozano1984automatic,tedrake2009lqr,konidaris2012robot,alatartsev2015robotic,eiband2021identification}. There are many ways to establish these transitions between skills or policies, and often different kinds of probabilistic methods are used to learn and smoothen the process. Thus, in this section, we survey the literature on different ways to represent such different behavior under different conditions for in-contact manipulation. 

A straightforward way, especially for tasks involving contact, is to use threshold values for the detected forces (for example, \cite{stolt2015detection}). The thresholds are sometimes referred to as \textit{guarded motions} in robotics \cite{will1975experimental,inoue1974force}. Johansson \etal have a considerable number of works regarding sequencing of skills based on detected contact wrench. They have been using transients \cite{stolt2015detection}, making the programming of such force-based segments easier for humans \cite{stolt2013robotic} 
and detecting successful assembly even without a \gls{ftsensor} in a snap-fit scenario \cite{karlsson2018detection}. There are also probabilistic methods used for such tasks with robots, such as the Bayesian Online Changepoint method \cite{su2016learning}.  A slightly more complex approach is to use thresholds in the contact formation space with suitable motion primitives based on RRT \cite{cheng2021contact}.  
 
\begin{figure}[tbp]
	\centering
	\includegraphics[width=0.6\columnwidth]{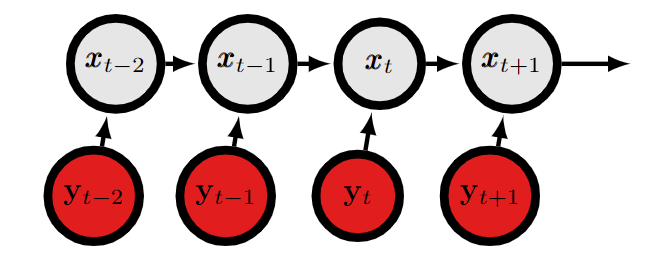}
	\caption{\change{A graphical illustration of a general HMM, where $x$ is the hidden state and $y$ the observation. The red nodes are observable and the gray ones hidden.}}
	\label{fig:hmm}
\end{figure}

\begin{figure}[tbp]
	\centering
	\includegraphics[width=0.6\columnwidth]{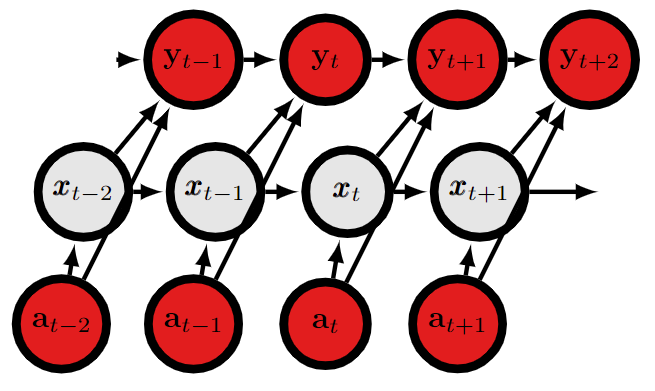}
	\caption{\change{An example of an autoregressive HMM from \cite{hagos2018}, where $x$ is the hidden state, $y$ the observation and $a$ an action, for example contact with the environment. The red nodes are observable and the gray ones hidden.}}
	\label{fig:tesfa_hmm}
\end{figure}

\change{There is a wide variety of probabilistic methods based upon the Markov process (meaning that a state $x_{t+1}$ only depends on the previous state $x_t$ and knowing the previous states gives no further predictive power). Since there is always uncertainty in the observations, a popular method to approach this problem is the \gls{hmm}, where it is assumed the state cannot be directly observed, but instead observations $y_t$ are made, which depend on $x_t$. This process is illustrated in Fig.~\ref{fig:hmm}.} The earliest works of using \gls{hmm} to segment a demonstration in \gls{lfd} (dividing a human demonstration automatically into sections that require different policies) and then choose a suitable policy during runtime were presented already in 1996 \cite{hovland1996skill}. A popular extension was to make the \gls{hmm} autoregressive with a beta process prior \cite{niekum2012learning,niekum2015learning,niekum2013incremental}. Whereas this model is not explicitly meant for in-contact tasks, further modifications were proposed that allowed the usage of the autoregressive state-space model together with an \gls{hmm} on tasks such as rotating a pepper mill \cite{kroemer2014learning,kroemer2015towards} and also for assembly tasks with an impedance controller requiring constant contact \cite{hagos2018}; \change{from here, an example of making the HMM autoregressive is shown in Fig.~\ref{fig:tesfa_hmm}}. Even further extension was shown to learn tasks requiring constant, varying contact in slicing and grating vegetables \cite{figueroa2016learning,figueroa2017transform}. 
Finally, sometimes it is useful to have time as a variable even in in-contact skills; \gls{hsmm} have been shown applicable to such cases \cite{racca2016learning} and can also be task-parametrized and used similarly as the motion primitives presented earlier \cite{le2021learning}. 

\begin{figure}[tbp]
	\centering
	\includegraphics[width=0.6\columnwidth]{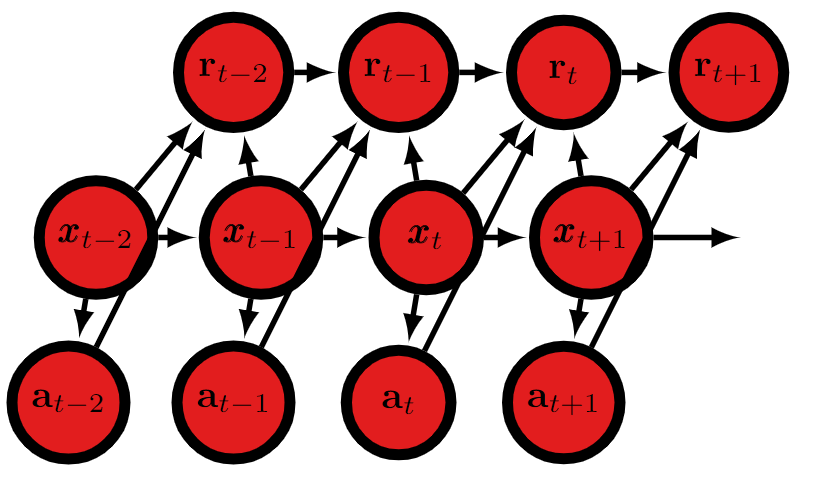}
	\caption{\change{Graphical illustration of an MDP, where $x$ is the (visible) state, $r$ the reward and $a$ the action.}}
	\label{fig:mdp}
\end{figure}

\change{When decisions and rewards are directly involved in the model, the Markov process becomes a  \textit{\gls{mdp}}, illustrated in Fig.~\ref{fig:mdp}.} \gls{mdp} and its extensions, most notably \gls{pomdp}, have been used extensively as representations for both planning \cite{guan2018efficient,calder2016wafr} and learning \cite{beltran2020learning,yuan2021multi} frameworks. Whereas both are very popular in robotics in general, \change{with different techniques available to reduce the notorious computational complexity of the \gls{pomdp}}, there is a limited number of in-contact tasks they have been used for; popular examples being \cite{Koval-ijrr2016,Sieverling2017}, who used POMDP to handle motions requiring both free space and in-contact motions. POMDP has also been used to leverage contact for localization both in classical tasks \cite{toussaint2014dual} and when teaching the robot to search for a goal from a human demonstration \cite{de2016learning,de2013learning,de2014learning}. 

An interesting use case especially for assembly in-contact tasks, which can be handled with many different methods, is \textit{exception strategies}; in difficult in-contact tasks the robot can often partially fail or end up with stuck workpieces, situations which can be managed through switching policies. For detecting such errors, for example HMM \cite{di2012hierarchical} and \gls{svm} \cite{rodriguez2010failure} have been used. Abu-dakka \etal \cite{abu2014solving} used random walk in case an assembly task failed and searching had to be done. Jasim, Plapper, and Voos \cite{jasim2014position} used an Archimedean spiral, which is guaranteed to find the goal with the correct resolution and starting position. {Chapter 5} in  \cite{kronander2015control} used incremental learning, where the human assists the robot during insertion if the robot gets stuck. Ehlers \etal \cite{ehlers2018search} proposed looking into the \textit{area coverage} literature instead of trajectory following to learn area or force-based search strategies; Shetty \etal \cite{shetty2021ergodic} employed ergodic control for similar search problems in peg-in-hole tasks. Hayami \etal \cite{hayami2021error} use the force signal on snapping assembly to figure out when and how things go wrong. Also, \cite{nemec2020learning} use \gls{pca} to detect context of assembly tasks to apply the correct exception strategy, and \cite{ortega2021dual} trained a \gls{nn} to detect errors in assembly with feedback from multiple \glspl{ftsensor}.

\section{Learning and planning of manipulation in contact}
\label{sec:learnplan}
When a suitable representation of a task is chosen, the actual policy, meaning usually the parameters of the representation, must be found for a skill the robot will then use to complete the task. This phase is typically called planning or learning. Even though many people associate their approach with either of the terms, there are often similarities, and nowadays combinations of these approaches; also several representations can be used with either approach. 

\change{The classical motion planning problem, which is at the core of either of the approaches, can be defined as follows: we have the robot's end-effector $E$ moving in the Cartesian workspace $W$ residing in $\Re^3$\footnote{May be a subspace, depending on the number of joints in the manipulator}. The manipulator can move the end-effector with an action $U$ according to a state transition function $f:X\times U \rightarrow X$. A plan $\pi$, essentially suitable parameters for one of the representations in Section~\ref{sec:representing}, must then be found, which applies actions $u\in U$ such that the end-effector moves from initial state $x_i \in W$ to a goal region $X_G \in W$, subject to a set of constraints: some of the constraints may be involved already in the definition of the workspace, such as joint limits or self-collisions.} 

\change{Whereas classically there would be an obstacle region to avoid, in the case of in-contact tasks the robot will specifically need to be in contact with the "obstacles"; this means that the plan cannot be only in Cartesian or joint space, but also the force or impedance profile of the robot must be represented.} There are two main approaches following the control methods presented in Section \ref{sec:control}: first, policies can consist of a sequence of force contacts, possibly together with motions, where the contacts are used for localization. Second, the policy can consist of motions with an impedance profile to provide the necessary compliance to overcome uncertainty in contact. In this section methods from both of these approaches are considered. Table~\ref{tab:learnplan} presents the surveyed papers divided into categories with respect to how the policy was eventually found, for those works where finding the policy was a major contribution.

In Section~\ref{sec:planning} we present planning approaches for manipulation in contact, and how contact can be leveraged for better or faster plans. In Section~\ref{sec:reinforcement} we present reinforcement learning approaches. Finally, we present imitation learning methods for in-contact manipulation in Section~\ref{sec:imitation}, where the idea is to use human expertise to find a suitable policy for the robot.

\begin{table*}[ht]
        \centering
        \small \caption{The papers divided across the methods of finding a suitable policy.}
        \vspace{-1mm}
        \begin{tabular}{@{}c|c @{}}
       Force control planning & \cite{jasim2014position,zhang2017peg, van2018comparative,Erdmann1988,stolt2012force,stolt2015detection, karlsson2018detection,yu1995fixture,yu1998complete,shao2020learning,almeida2016folding,stolt2015robotic,park2020compliant}. \\ \hline
       Contact for localization & \cite{pall2018contingent,newman2001interpretation,gadeyne2005bayesian,lefebvre2005online,lefebvre2005polyhedral,shao2020learning,baum2017opening} \\ \hline
       Compliant motion planning & \cite{calder2016wafr,Koval-ijrr2016,Sieverling2017,Ji2001,guan2018efficient,Chavan-Dafle2020,wirnshofer2018robust,nottensteiner2020robust} \\ \hline
       LfD & \cite{steinmetz2015simultaneous, montebelli2015handing, hazara2016reinforcement, nemec2018learning, deng2016learning,gams2016adaptation,kramberger2020adapting,kramberger2018passivity,gams2014coupling,abu2014solving,abu2015adaptation} \\ {} & \cite{abu2018force,denivsa2016learning,kramberger2016transfer,kramberger2017generalization,petrivc2018accelerated,tykal2016incrementally,figueroa2016learning, ureche2018constraints,khoramshahi2020arm,rozo2015learning} \\ \hline
       RL & \cite{valassakis2020crossing,kalakrishnan2011learning,hazara2016reinforcement,levine2015learning,inoue2017deep,hou2020fuzzy,kuo2021uncertainty,hoppe2019planning,wang2021hybrid,kaspar2020sim} \\ {} & \cite{apolinarska2021robotic,beltran2020learning,schoettler2020meta,bogdanovic2020learning,oikawa2021reinforcement,martin2019variable,kulkarni2021learning,luo2019reinforcement,belousov2022robotic}. \\ \hline
  \end{tabular}
  \label{tab:learnplan}
\end{table*}


\subsection{Planning of manipulation in contact}
\label{sec:planning}

\change{As compliant motions require constant interaction with the environment, we consider the kind of kinematic motion planning defined earlier, and not symbolic, higher-level task planning.} The origins of motion planning for compliant motions can be traced to fine motion planning \cite{finemotionplans,backprojections}; however, this approach always suffered from a high computational load, as did another early approach for planning compliant motions, preimage planning 
\cite{lozano1984automatic}. Another early approach was the use of threshold values to plan for assembly tasks, such as \cite{schimmels1991force}, and there have been attempts to use other popular methods, such as the probabilistic roadmaps, on in-contact skills \cite{Simeon2004}. Additionally, whereas most of the methods reported in this section have been proven on small-scale tasks, also excavation has been successfully done with planning \cite{jud2017planning}.

The use of force values for planning is perhaps more popular than compliance. For peg-in-hole assembly, there are several planning methods taking advantage of contacts: whereas the easier approach is to attach a force sensor to the robot arm \cite{jasim2014position,zhang2017peg, van2018comparative}, it is also possible to plan force-based assembly tasks without a force sensor \cite{Erdmann1988,stolt2012force}. Sometimes these threshold values are also called transients \cite{stolt2015detection, karlsson2018detection}. Another popular term is fixture loading; fixtures are physical guides that help guide an assembly task, where planning of motions has been shown useful \cite{yu1995fixture,yu1998complete}. There is also novel development in this field, such as finding optimal placements for fixtures so that they can mitigate uncertainty  \cite{shao2020learning}. Further, planning has also been used on assembly tasks where an accurate combination of force and motion is required to accomplish a task, such as folding \cite{almeida2016folding} or snapping \cite{stolt2015robotic}. Also, general force profiles can be used to plan compliant motions with the use of force control \cite{park2020compliant}.

Even though planning optimal compliant motions is infeasible, there are many approximations through which near-optimal solutions can be found. Phillips-Grafflin and Berenson \cite{calder2016wafr} used an \gls{mdp} approach to overcome actuation 
uncertainties for planning assembly tasks taking advantage of compliance. Also, the extension of imperfect information, \gls{pomdp} has been used as the representation for planning \cite{Koval-ijrr2016}, targeting pushing with a switching policy depending on the contact state, or for completely interleaving in-contact and free space motions \cite{Sieverling2017}. Additionally, the earlier mentioned probabilistic roadmaps have also been used together with contact states to plan compliant motions \cite{Ji2001}. A notable result in this area was achieved by  Guan, Vega-Brown, and Roy \cite{guan2018efficient}, who showed near-optimal planning of compliant motions using a feedback \gls{mdp}. The often complex planning problem can also be simplified by realizing that in compliant motions, different motions of the robot can result in the same kind of motion of the object \cite{Chavan-Dafle2020}. Finally, when rich information about the workpieces is known beforehand, such as the CAD model \cite{wirnshofer2018robust} or the mesh model \cite{nottensteiner2020robust}, planning can be done efficiently. 

An interesting approach is to use contacts as localization to mitigate uncertainty; the straightforward approach is to consider point-contacts for localization \cite{pall2018contingent}. An approach beyond contact thresholds is the concept of \gls{cs}, which means finding the exact contact point, or points, of a tool. This can be done with only a force-position mapping to solve the task, in this case peg-in-hole \cite{newman2001interpretation}, but there is also a wide literature on detecting where contact to an assemblable piece occurs and how to make a more general plan: first, the state of contact (for example, point contact or face-to-face contact) needs to be detected \cite{gadeyne2005bayesian} before they can be exploited during autonomous compliant motions by the help of Kalman filtering and Bayesian estimation \cite{lefebvre2005online,lefebvre2005polyhedral}. To improve such strategies, \cite{shao2020learning} learn how to place fixtures in the environment to optimize this kind of uncertainty mitigation offered by contact. Finally, physical exploration can also be required to perform complex and new tasks, which was addressed in \cite{baum2017opening}.


\subsection{Reinforcement learning of manipulation in contact}
\label{sec:reinforcement}
Reinforcement learning (RL) methods aim to learn an optimal control policy for stochastic time-series systems by exploring the dynamics of the environment \cite{sutton2018reinforcement}. 
Reinforcement learning has recently seen wide interest in robotics, with multiple surveys spanning the entire spectrum of RL including \cite{kober2013reinforcement}. 
In this section, we focus solely on the use of reinforcement learning for in-contact manipulation tasks.
Types of tasks that have been addressed using RL include pushing \cite{valassakis2020crossing}, manipulation of articulated objects such as door opening \cite{kalakrishnan2011learning}, tool use \cite{hazara2016reinforcement}, as well as peg-in-hole and similar assembly tasks \cite{levine2015learning,inoue2017deep,hou2020fuzzy}. 

Learning in-contact manipulation directly in a physical robot system using reinforcement learning is challenging because the exploration needed for the learning is often a safety hazard due to potentially high contact forces in high stiffness environment interactions. 
Methods exploring physical systems alleviate the potential hazard by limiting contact forces using torque control (e.g.~\cite{levine2015learning}), impedance control with limited stiffness (e.g.~\cite{hazara2016reinforcement}), or explicit limiting of actions using safety based constraints \cite{kuo2021uncertainty,liu2022robot}.
The data-hungry nature of reinforcement learning presents an additional challenge for learning directly in a physical system. 
This is often alleviated by using a human demonstration as a starting point for the RL-based policy optimization \cite{hazara2016reinforcement, hoppe2019planning,wang2021hybrid}. 

In contrast to learning in a physical system, policies can also be learned in a simulation environment before executing them in the physical system. 
This allows the exploration to be performed safely in simulation and provides access to vast amounts of training data.
However, the approach results in another challenge in the form of the reality gap between the simulation and the physical system. 
The gap can be reduced by calibrating the simulation using system identification \cite{kaspar2020sim}.
Alternatively, policies can be robustified by training them with simulated noise \cite{valassakis2020crossing,apolinarska2021robotic}, or over a known distribution of simulator parameters also known as domain randomization \cite{beltran2020learning}. Also, a simulation-trained policy may be further refined in real-world, for example using meta-learning to learn adaptable policies \cite{schoettler2020meta,arndt2020meta}. \change{Finally, there are ongoing research efforts for simulators to aid the process of using RL-methods for in-contact skills: whereas Mujoco \cite{todorov2012mujoco} is still popular even after 10 years, there are newer, often more specialized simulation tools: \emph{RLBench} for a multitude of simple or multistage tasks \cite{james2020rlbench}, \emph{Meta-world} for evaluating manipulation tasks learned with meta-RL and multitask learning \cite{yu2020meta}, \emph{SAPIEN} for home tasks including articulated motions \cite{xiang2020sapien}, and \emph{ReForm} for linear, deformable objects \cite{laezza2021reform}. Simulators are also used to evaluate the capability of a method to cross the reality gap, to provide repeatable substitutes for real-world experiments. However, the actual capability can only be evaluated using physical real-world experiments, since simulated reality gaps differ from the physical ones, and because pre-recorded data sets cannot be used to evaluate the performance of a closed loop system.}

RL has been used to optimize various aspects of in-contact policies, including controller set point trajectories (assuming a known controller), controller parameters (assuming known target trajectories and controller structure), and entire feedback policies. 
Set point trajectories can be learned e.g.\ for end-effector poses of an impedance-controlled manipulator \cite{hazara2016reinforcement,bogdanovic2020learning}.
Another common approach is to use RL to optimize control parameters such as gains \cite{beltran2020learning} or compliance parameters \cite{bogdanovic2020learning,oikawa2021reinforcement,wang2021hybrid}. 
In the case of feedback policies, the choice of action space (e.g.\ velocity vs force) has been found to be application dependent \cite{martin2019variable}.
RL can also be combined with other control and decision making approaches, including classical control methods such as impedance control \cite{kulkarni2021learning} and operational space control \cite{luo2019reinforcement}, planning \cite{belousov2022robotic}, or fuzzy logic \cite{hou2020fuzzy}.

\subsection{Learning from demonstration of manipulation in contact}
\label{sec:imitation}
In \gls{lfd}, also known as imitation learning or \gls{pbd}, a user shows the robot how a task should be done. We note that many works of \gls{lfd} for in-contact tasks focus on finding novel, better representations that can better encode the task. We covered already in Section~\ref{sec:representing} these parts, such as extending popular approaches (DMP,GMM,SEDS) to in-contact tasks, as well as less-known representations meant for in-contact tasks. Thus, in this section, we will focus more on issues such as good demonstration, force data gathering, and number of required demonstrations. 

Whereas \gls{lfd} is already a well-established idea in robotics \cite{ravichandar2020recent}, there is a specific requirement to perform \gls{lfd} with in-contact tasks: the contact force during the demonstration must be either measured or estimated. This is a crucial part when deciding on the demonstration method: with perhaps the most popular demonstration method, grabbing the robot and leading it through the motions (also known as kinesthetic teaching), a \gls{ftsensor} is often required since the joint torque sensors cannot observe the force if the user holds the robot near the tool (there are however various ways to place the sensor, with two different ones presented in \cite{brunete2016user} and \cite{montebelli2015handing}). For teleoperation with electrically actuated robots, motor currents, motor output torques \cite{stolt2012force} or torque sensors at the joints that robots such as the Franka Emika Panda have, can be used to estimate the contact force during the demonstration; for hydraulically actuated robots, hydraulic fluid chamber pressures can be used to estimate the contact forces \cite{koivumaki2015stability}. Other proven methods include handheld devices with methods to modulate impedance \cite{peternel2015human,peternel2018robotic,kronander2012online} or electromyography sensors on the demonstrator \cite{peternel2017method}. There are also certain things a human does, such as slipping the grip, which need special methods for a robot to detect in a demonstration \cite{hagenow2021recognizing}.

Once the force data is gathered, there are various options to encode the data, most of which were handled already in Section~\ref{sec:representing}. The requirements may also vary: \gls{dmp} and its variants can be used to learn from a single demonstration only, but if multiple demonstrations are used they need to be aligned in time, \change{using, for example, \gls{dtw} (but also simpler methods have been proposed, for example, \cite{pervez2017learning}). Also, the number of "primitives", essentially meaning the number of the forcing functions, must be decided beforehand; however, this is typical for most of the representations, such as the number of Gaussians for the GMM. The \gls{gmm}-based methods can naturally handle multiple demonstrations, but often require more than one to start with; also the GMM-based methods employing dynamical systems are time-invariant and can be proven to be stable, unlike DMP-based methods.}
These representations, and their extensions, are the most popular ones used in \gls{lfd} and are covered in more depth already in Section~\ref{sec:representing}, with both continuous and discrete representations being used, often hierarchically. In addition, methods mostly used in planning have also been adapted to \gls{lfd}, such as contact state estimation \cite{meeussen2007contact}. A common problem with many of the aforementioned strategies is that the segmentation part often produces too many segments; pruning and joining of these segments to simplify execution, regardless of the representation, has been suggested in \cite{lioutikov2015probabilistic,lioutikov2017learning}.



Another challenge in \gls{lfd} with in-contact tasks is that often the robot should exploit the environment for more accurate positioning; however, as users often give as good demonstrations as they can, it can be challenging to gather demonstrations where uncertainty forces the demonstrator to exploit the environment. There are several works looking into how to give good demonstrations, which show that explicit instructions help the demonstrator \cite{sena2018teaching,sena2019quantifying}. More implicit methods include incremental learning \cite{tykal2016incrementally,dimeas2020progressive}, where the robot gradually develops the correct behavior from multiple demonstrations and corrections, and active learning \cite{chao2010transparent} where the robot communicates to the teacher what sort of demonstrations are required. Whereas there are recent works about active learning for manipulation \cite{bestick2018learning,maeda2017active} for in-contact tasks there are currently no known active learning approaches. Finally, the task can be made more difficult for the demonstrator to observe the required behavior, for example by using blurring glasses \cite{eppner2015exploitation} or simply blindfolding the teacher \cite{de2016learning,ehlers2018search}.

Finally, it is an interesting question whether \gls{lfd} is even a good way to learn such tasks, or are the human demonstrations already sub-optimal. Whereas there is no clear answer, at least for the classic peg-in-hole task people do have an efficient strategy to learn from \cite{savarimuthu2013analysis}. Whereas \gls{lfd} can often be used to quickly learn tasks in, for example, assembly lines, when more optimization is required a powerful strategy is pairing \gls{lfd} with \gls{rl}, such that the robot's policy is further optimized offline with reinforcement learning or optimization methods \cite{hazara2016reinforcement,roveda2020assembly,wang2021hybrid}, or the demonstrations can be used to enhance the effectiveness of classical motion planning methods \cite{guo2021geometric}. 


\section{Conclusions and future work}
\label{sec:conclusions}
In this survey, we provide readers with a comprehensive look into various state-of-the-art methods for making robots perform in-contact tasks, which are a natural next step for increasing the automation level in factories and at homes. After laying out the kind of in-contact tasks that have been performed by robots, we further showed how to control robots during such tasks, followed by representations used to encode such tasks and finally methods to exploit the representations in a correct way.

In-contact tasks are very often required both in factory assembly and at service tasks. For factories, leveraging compliance can further increase the number of tasks that can be performed by robots, also for cases where the products often vary; at homes, restaurants, and other such environments, mastering in-contact tasks will endow service robots of the future with an increasing number of abilities. Whereas contact between the robot and the environment is not any more viewed as a nuisance, there is still a long way to view it as useful as people do; thus, we hope that this survey will help people delve into the area of in-contact tasks, exploit the researched ones and come up with new ideas.

Several areas for future work can be identified from this survey. 
\change{Currently, most of the tasks are completed with stiff objects; however, there are only few works ( \cite{wirnshofer2018robust,laezza2021reform}) with objects that bend. Whereas cloth folding is a researched topic (for example \cite{tsurumine2019deep,petrik2020static}), assembly tasks with objects that bend is even more complex than that, since the interplay of deformation and friction creates a complex physics task to reason about. Whereas handling of exception strategies has been slowly getting traction (see Table~\ref{tab:tasks}), there is a lot of room for improvement, since this kind of behavior is where people still often overcome robots, especially in industrial use cases; if a robot needs to be manually handled every time a non-typical error appears, it is not very beneficial for the industry. Another interesting in-contact task that provides unique challenges not yet tackled is painting;} whereas similar to wiping, painting a wall such that the result is not uneven requires a very delicate manipulation of forces, which would require a lot from a physics engine to simulate, or very accurate recording of motions for LfD, and naturally the ability to repeat these forces by the controller. Also, there are still certain assembly tasks even with stiff objects, such as board-to-board matings, where the accuracy can be improved by leveraging contact. 

\change{Besides teaching robots new tasks, the learning time and generalization abilities must be improved.} Teaching new in-contact tasks to robots has to be fast; methods such as one-shot imitation learning and transfer learning can help in this, when they are easy enough to deploy in factories and also capable of learning even complex tasks from a single demonstration. Also, the improvements in physics simulation can allow better transfer learnings, such that even difficult tasks such as exploiting contact and friction can be learned in simulation before being deployed to the real world; \change{there are surprisingly few works even using domain randomization on dynamics \cite{peng2018sim} for in-contact tasks, even though the topic has been popular elsewhere in robotics. However, as domain randomization is slow, having accurate enough simulators to avoid it altogether would be beneficial \cite{tobin2017domain})}. For simulations, also demonstrations can be given in \gls{vr} with sufficiently good physics engines, something that is under active development \cite{dyrstad2018teaching,zhang2018deep}. Another \gls{lfd}-related future work is the \gls{hri} side; how would a non-roboticist understand what is a good demonstration for the robot to learn various tricks in difficult tasks, such as wood handling? Finally, even though there is research on inferring contact forces with vision only \cite{pham2015towards,pham2017hand}, this work is limited to grasping; with novel deep learning methodologies, it should be possible for robots to learn tasks even such as sawing from only watching a video. 

\bibliographystyle{ieeetr} 
\bibliography{biblio}

\end{document}